%% file: arXiv_ICML.tex
\documentclass{article}

\usepackage{microtype}
\usepackage{graphicx}
\usepackage{subfigure}
\usepackage{booktabs} 

\usepackage{hyperref}

\usepackage[arxiv]{icml2024}

\usepackage{amsmath}
\usepackage{amssymb}
\usepackage{mathtools}
\usepackage{amsthm}

\usepackage[capitalize,noabbrev]{cleveref}

\theoremstyle{plain}

\theoremstyle{definition}

\theoremstyle{remark}

\usepackage[textsize=tiny]{todonotes}

\usepackage{multirow}
\usepackage{tabularx}
\usepackage{ragged2e}
\usepackage{makecell}

\usepackage{bm}

\newcommand{\bxV}{\bm{x}_{\textsc{V}}}
\newcommand{\bxI}{\bm{x}_{\textsc{I}}}
\newcommand{\bxR}{\bm{x}_{\textsc{R}}}

\newcommand{\psiV}{\psi_{\textsc{V}}}
\newcommand{\psiL}{\psi_{\textsc{L}}}
\newcommand{\psiC}{\psi_{\textsc{C}}}

\newcommand{\bthetaV}{\bm{\theta}_{\textsc{V}}}
\newcommand{\bthetaL}{\bm{\theta}_{\textsc{L}}}

\newcommand{\bthetaC}{\bm{\theta}_{\textsc{C}}}

\DeclareMathOperator*{\argmin}{arg\,min}

\icmltitlerunning{SVIT: Scaling up Visual Instruction Tuning}

\begin{document}

\twocolumn[
\icmltitle{SVIT: Scaling up Visual Instruction Tuning}



\icmlsetsymbol{equal}{*}

\begin{icmlauthorlist}
\icmlauthor{Bo Zhao}{equal,baai}
\icmlauthor{Boya Wu}{equal,baai}
\icmlauthor{Muyang He}{equal,baai,pku} 
\icmlauthor{Tiejun Huang}{baai,pku}
\end{icmlauthorlist}

\icmlaffiliation{baai}{Beijing Academy of Artificial Intelligence, Beijing, China}
\icmlaffiliation{pku}{Peking University, Beijing, China}

\icmlcorrespondingauthor{Bo Zhao}{zhaobo@baai.ac.cn}

\icmlkeywords{Machine Learning, ICML}

\vskip 0.3in
]



\printAffiliationsAndNotice{\icmlEqualContribution} 

\begin{abstract}
Thanks to the emerging of foundation models, the large language and vision models are integrated to acquire the multimodal ability of visual captioning, question answering, etc. Although existing multimodal models present impressive performance of visual understanding and reasoning, their limits are still largely under-explored due to the scarcity of high-quality instruction tuning data. To push the limits of multimodal capability, we Scale up Visual Instruction Tuning (SVIT) by constructing a dataset of 4.2 million visual instruction tuning data including 1.6M conversation question-answer (QA) pairs, 1.6M complex reasoning QA pairs, 1.0M referring QA pairs and 106K detailed image descriptions. Besides the volume, the proposed dataset is also featured by the high quality and rich diversity, which is generated by prompting GPT-4 with the abundant manual annotations of images. We also propose a new data recipe to select subset with better diversity and balance, which evokes model's superior capabilities. Extensive experiments verify that SVIT-v1.5, trained on the proposed dataset, outperforms state-of-the-art Multimodal Large Language Models on popular benchmarks. The data and code are publicly available at \url{https://github.com/BAAI-DCAI/Visual-Instruction-Tuning}.
\end{abstract}

\input{introduction}
\input{relatedwork}

\input{dataset}

\input{method}
\input{experiments}
\input{conclusion}

\section*{Acknowledgment}
This work is funded by the following grants: National Key R\&D Program of China (2021ZD0111102) and NSFC-62306046.

\bibliography{refs}
\bibliographystyle{icml2024}

\newpage
\appendix
\onecolumn
\input{supplementary}

\end{document}

%% file: introduction.tex
\section{Introduction}
The great success of large language models (LLMs), e.g. BERT  \cite{BERT}, T5 \cite{T5}, GPT-2 \cite{GPT2}, GPT-3 \cite{GPT3}, have motivated the advancement of vision  \cite{ViT, SwinT, MAE} and multimodality  \cite{CLIP, Flamingo, MiniGPT4, LLaVA} in terms of architecture design and learning paradigm.
Recently, GPT-4  \cite{GPT-4} demonstrates impressive multimodal understanding and reasoning abilities, accepting image and text inputs. Inspired by GPT-4, Multimodal Large Language Models (MLLMs) bridging language and vision models have achieved remarkable progress in multiple visual understanding and reasoning tasks, e.g. visual captioning \cite{Blip-2}, dialogue \cite{Flamingo} and question answering \cite{MiniGPT4, LLaVA}. 

Typically, the multimodal models are pre-trained on large multimodal datasets, e.g. LAION-2B \cite{LAION5B}, CC-12M \cite{CC12M}, YFCC-100M \cite{YFCC100M} and MMC4 \cite{MMC4}, that contain millions to billions roughly-aligned image-text pairs from the web. Then, precise vision-language data pairs are used to finetune the models. Like the success of language instruction tuning, visual instruction tuning has become the key to the multimodal performance.
However, due to the high construction cost, existing visual instruction datasets are still in small scale and less informative. 
Several works convert the image captioning and VQA datasets \cite{MSCOCO, VQA2015, GQA2019, goyal2017making} into instruction tuning data by manually adding a few instructions  \cite{instructblip}. However, these captions and questions/answers are usually short and focus on visual perception and simple questions, which may lead to ineffective model training \cite{Multimodal-gpt}. To generate more informative visual instruction data, GPTs are introduced. LLaVA \cite{LLaVA} contributes a large visual instruction dataset containing {158K} data by prompting GPT-4 with five captions and a few object bounding boxes associated with images from COCO dataset \cite{MSCOCO}. Meanwhile, MiniGPT-4 \cite{MiniGPT4} creates 3,500 image-text pairs by refining model's output using ChatGPT. The language-only GPT models have difficulty in precisely imagining the whole picture from the limited input. Thus, the generated instruction tuning data lacks diversity and complexity.

\begin{figure*}
  \centering
  \includegraphics[width=\linewidth]{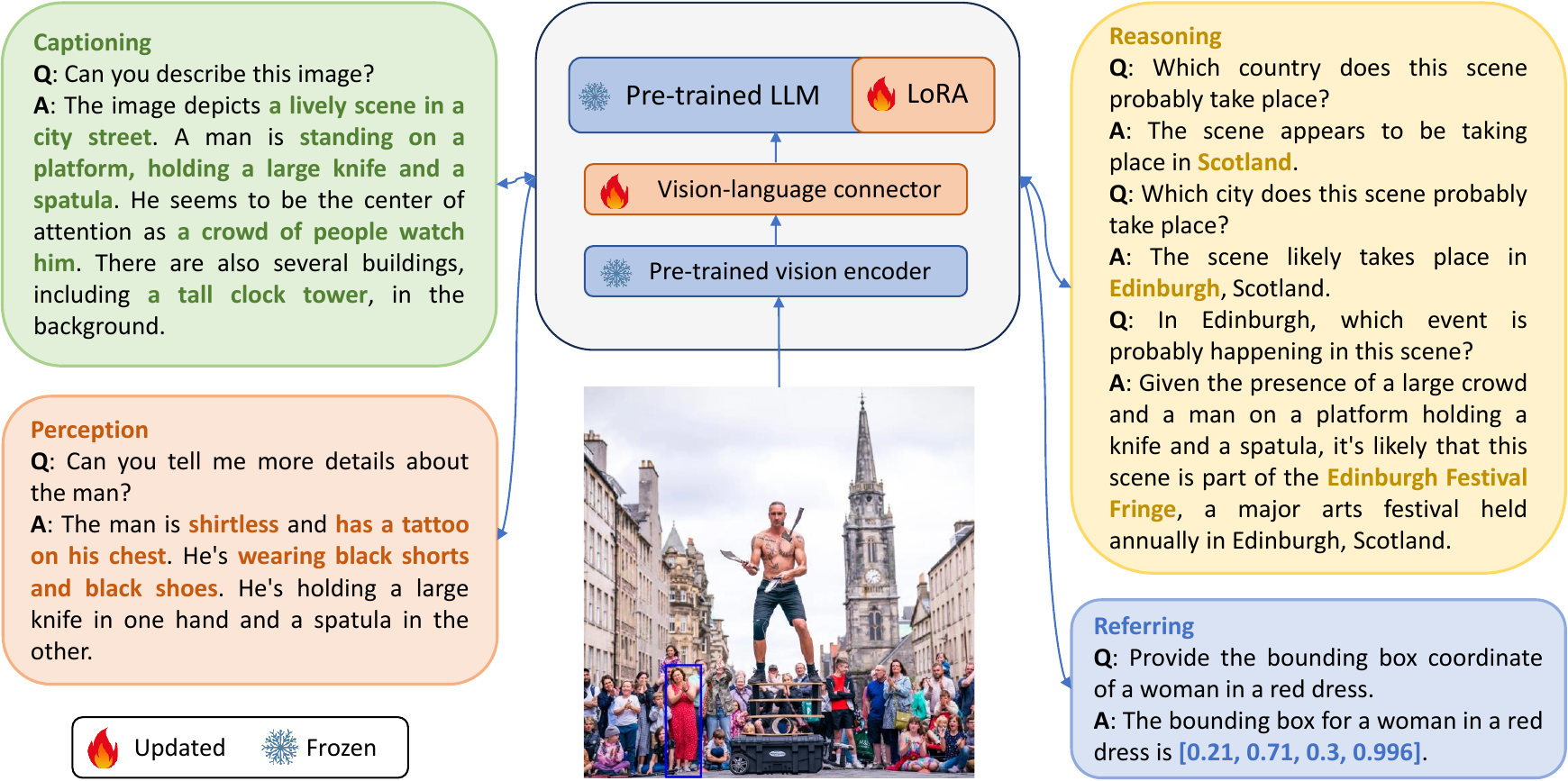}
  \caption{{SVIT-v1.5 (LoRA)} model architecture and abilities.}
  \label{fig:model}
\end{figure*}

To push the limits of large multimodal models, we Scale up Visual Instruction Tuning (SVIT) and propose a large-scale dataset with 4.2 million informative instruction tuning data, including 1.6M conversation QA pairs, 1.6M complex reasoning QA pairs, 1.0M referring QA pairs and 106K detailed descriptions. \cref{tab:datasets} shows that SVIT is $20\times$ larger than LLaVA dataset. 
To enrich the diversity and informativeness of instruction tuning data, we construct SVIT based on Visual Genome \cite{VisualGenome} which has abundant manual annotations and GPT-4 which has the best multimodal capability. We prompt the language-only GPT-4 ChatBot with image-level descriptions, detailed region descriptions and object bounding boxes. 
We further study the data efficiency and propose a new data recipe that outputs subset with better diversity and balance. 
Then, a more powerful model, SVIT-v1.5, is trained on the proposed dataset, as illustrated in \cref{fig:model}.
Extensive experiments verify that our model reveals impressive ability in visual perception and reasoning, and achieves noticeable performance improvements over the state of the art.

We summarize the main contributions of this paper:
\begin{enumerate}
    \item We present 4.2M high-quality instruction data of 1.6M conversation QA pairs, 1.6M complex reasoning QA pairs, 1.0M referring QA pairs and 106K detailed image descriptions.
    \item We propose a new data recipe that selects an informative subset of diverse and balanced training data to better match the downstream tasks. 
    \item We scale up visual instruction tuning and contribute a better model -- SVIT-v1.5 that outperforms state-of-the-art MLLMs including {LLaVA-v1.5, Qwen-VL-Chat and InstructBLIP on popular benchmarks}.
\end{enumerate}

\begin{table*}[ht]
\small
\vspace{-5pt}
  \caption{Comparing SVIT to similar vision-language instruction datasets generated by GPT. { $^*$LLaVAR collects 422K noisy instruction-following data using OCR results and 16K high-quality data using GPT-4.}}
  \vskip 0.15in
  \label{tab:datasets}
  \centering
  \begin{tabular}{l|cccc|cc|c}
    \toprule
    \multirow{2}{*}{Dataset}  & \multirow{2}{*}{\#Image}   & \#Object      & \#Region & \#Image & \#Instruction & \#Response & \multirow{2}{*}{GPT} \\
    & & BBox & Description  & Caption & Question & Answer & \\
    \midrule
    MiniGPT-4       & 3.5K      & -         & -         & -         & 4         & 3.5K      & GPT-3.5   \\
    LLaVAR$^*$      & 16K       & -         & -         & -         & 16K       & 16K       & GPT-4     \\ 
    LLaVA           & 81.5K     & 600K      & -         & 404.7K    & {158K}      & 158K     & GPT-4     \\
    SVIT            & {108.1K}  & {3.8M}    & {5.4M}    & {257.6K}  & {4.2M}    & {4.2M}    & GPT-4   \\
    \bottomrule
  \end{tabular}
\end{table*}


%% file: relatedwork.tex
\section{Related Work}
\subsection{Multimodal Models} 
Existing multimodal solutions can be roughly split into two categories: 1) multimodal systems, e.g. Visual ChatGPT \cite{VisualGPT}, X-Decoder \cite{XDecoder} and InternGPT \cite{InternGPT}, in which multiple language and vision models are coordinated by a LLM manager/controller to deal with different tasks, 2) end-to-end differentiable multimodal models, e.g. Flamingo \cite{Flamingo}, BLIP-2 \cite{Blip-2}, Kosmos \cite{KOSMOS-1, KOSMOS-2}, MiniGPT-4 \cite{MiniGPT4}, LLaVA \cite{LLaVA}, InstructBLIP \cite{instructblip} which input both vision and language tokens into LLM. In this paper, we focus on the end-to-end differentiable multimodal models, which are lightweight and concise for research.

The end-to-end multimodal models contain pre-trained vision and language models and a learnable module to fuse both. Flamingo \cite{Flamingo} learns gated cross-attention layers to condition the frozen LLM on visual tokens, demonstrating excellent in-context few-shot learning performance. \citet{Blip-2} design Q-Former to bridge the image encoder and LLM in a two-stage training strategy, which shows emerging capability of zero-shot instructed image-to-text generation. 
By leveraging advanced LLMs, i.e. LLaMA \cite{Llama} and Vicuna \cite{Vicuna}, multimodal models LLaVA \cite{LLaVA} and MiniGPT-4 \cite{MiniGPT4} are built by transforming visual tokens to language tokens with only one linear layer, while InstructBLIP \cite{instructblip} learns a Q-Former to bridge vison and language models.

\subsection{Multimodal Instruction Tuning}
The success of multimodal models, e.g. LLaVA, MiniGPT-4 and InstructBLIP, relies on the high-quality image-text data for finetuning models, which is named visual instruction tuning in \citet{LLaVA}. Previous work \cite{Multimodal-gpt} finds that simply constructing training set based on existing VQA datasets \cite{goyal2017making, GQA2019} with short answers will degrade the model performance. 
To boost the performance, \citet{MiniGPT4} collect 3,500 high-quality image-text pairs by refining their model's outputs using ChatGPT. More natural and reliable responses are produced by finetuning the model on the refined data.
\citet{LLaVA} for the first time systematically construct a large visual instruction tuning dataset -- LLaVA-Instruct-150K. They prompt GPT-4 to generate questions and answers by feeding it image-level captions and object bounding boxes of each image from COCO dataset \cite{MSCOCO}.
To better understand text-rich images, \citet{LLaVAR} present LLaVAR that collects 422K noisy instruction-following data using OCR results and 16K high-quality data using GPT-4.
\citet{instructblip} collect 26 public datasets including LLaVA-Instruct-150K to construct visual instruction tuning data. However, most of these public datasets contain short questions and answers that focus on visual perception.
\citet{M3IT} build M$^3$IT by converting 40 datasets into a unified vision-to-text schema. They utilize ChatGPT to paraphrase the short answers in original VQA datasets.
Beyond above works, we prompt the powerful GPT-4 with rich annotations of image-level captions, region-level descriptions and object bounding boxes that are from Visual Genome \cite{VisualGenome} and COCO dataset \cite{MSCOCO}. The generated 4.2M visual instruction data cover diverse tasks of visual perception, reasoning and planing. 

There are also some works that contribute multimodal instruction data of videos \cite{MIMIC-IT}, RGB-D images \cite{MIMIC-IT}, speech \cite{SpeechGPT}, audio \cite{zhang2023video}, etc. For instance, EgoCOT \cite{EmbodiedGPT} prompts ChatGPT with video captions to generate instructions and responses of detailed embodied planning. MIMIC-IT \cite{MIMIC-IT} collects visual data from multiple datasets, and prompts ChatGPT to generate instruction-response pairs. Most of its data are constructed based on the egocentric videos from E4D dataset \cite{Ego4d2022}.  








%% file: dataset.tex
\section{Dataset Construction}
\subsection{Source Data}
We build SVIT based on Visual Genome \cite{VisualGenome} dataset that comprises 108,077 images with dense annotations within each image, including region descriptions, objects, attributes, relationships, etc. Since Visual Genome is partially sourced from COCO dataset \cite{MSCOCO}, we also collect captions for images from COCO dataset. Generally, each image in COCO dataset has 5 captions, focusing on the high-level appearance. As an image usually contains rich objects and regions that cannot be completely described in a general caption, Visual Genome serves as a valuable source, offering abundant annotations of the visual details. On average, Visual Genome provides 42 human-generated region descriptions and 21 objects per image, with each region and object located by a bounding box.
Leveraging these annotations, we are able to gather thorough and detailed descriptions for all images, which are made up of three key components: (1) the 257,633 captions from COCO dataset; (2) the 3,802,374 object names and their corresponding bounding boxes from Visual Genome; (3) the 5,406,592 region descriptions and their corresponding bounding boxes from Visual Genome. 

\begin{figure*}
  \centering
  \includegraphics[width=0.9\textwidth]{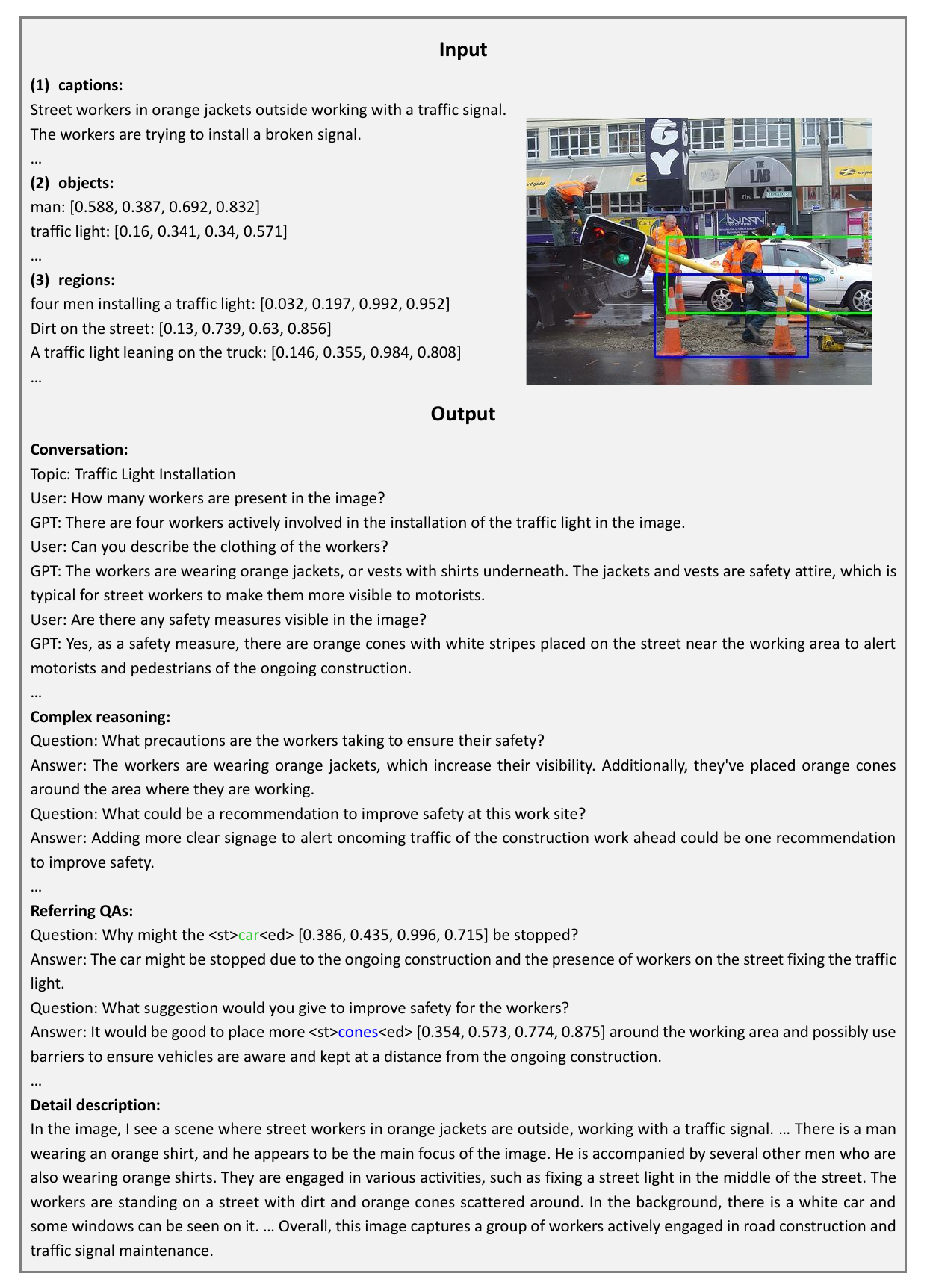}
  \caption{The example input to GPT-4 and the responses for three tasks. Note that the image is only shown here for reference and not provided to GPT-4. The colored phrases in referring QAs correspond with bounding boxes of that color in the image.}
  \label{fig:example}
\end{figure*}

\subsection{Instruction Data Generation}
Inspired by LLaVA \cite{LLaVA}, we design four tasks and prompt the language-only GPT-4 ChatBot to generate the questions and answers accordingly. The prompts are summarized in \cref{fig:prompts} and \cref{fig:region-prompt} in the Appendix. Since GPT-4 demonstrates excellent performance even with zero-shot learning, we do not provide any examples for GPT-4 in order to encourage the innovation and diversity of the generated contents.

\begin{itemize}
  \item \textbf{Conversation.} We prompt GPT-4 to design 3 conversations between a person and GPT-4 talking about the image. Each conversation should include 5 question and answer pairs (QAs). The content of the conversation should be logically connected. GPT-4 thinks about the topic first and then generates the conversation according to the topic. The topics can be about the visual perception, reasoning, event planning, etc. 
  \item \textbf{Complex reasoning.} 15 complex reasoning QAs about each image are generated using GPT-4. The questions can be asking why things happen that way, suggestions to the people in the image, etc. When providing the answer to a complex question, we prompt GPT-4 to think step by step and include reasoning details in the answer.
  \item \textbf{Referring QAs.} We prompt GPT-4 to create 10 question and answer pairs of specific regions in the image. When referring to any object in the question or answer, always wrap it with prefix ``$<$st$>$'', suffix ``$<$ed$>$'' and attach its normalized bounding box after it, in the format of ``$<$st$>$object$<$ed$>$ [x1, y1, x2, y2]''. If multiple objects are referred to, attach all the corresponding bounding boxes after them, e.g., ``$<$st$>$objects$<$ed$>$ [x1, y1, x2, y2], [x1, y1, x2, y2]''.
  \item \textbf{Detail description.} We use GPT-4 to describe the image in detail. The description may include appearances, actions, the number of objects, object positions, background details, etc.
\end{itemize}

\begin{figure*}
  \centering
  \includegraphics[width=0.9\textwidth]{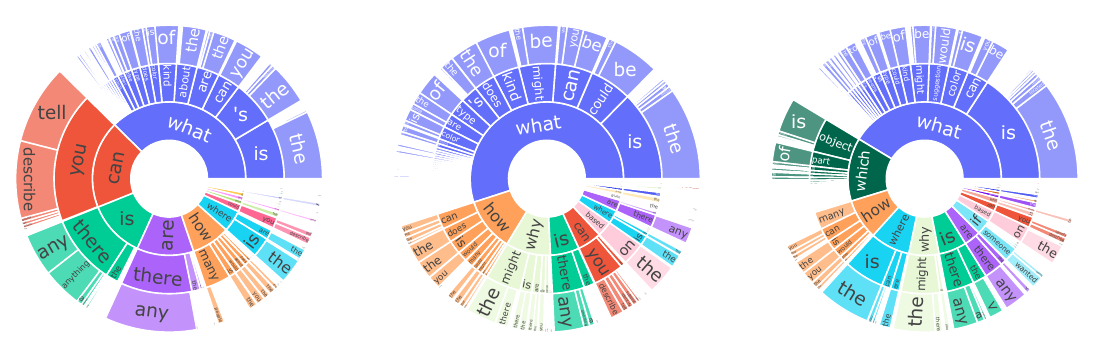}
  \caption{The distribution of question types in \emph{conversations} (left), \emph{complex reasoning} (middle), \emph{referring QAs} (right) by the first three words. The angle of each sector represents the proportion of each category.}
  \label{fig:distribution}
\end{figure*}

\cref{fig:example} illustrates an example input and the GPT-4 output for each task. For rich diversity, we further randomly sample an instruction for detail description task, e.g., ``can you describe the image in detail''. The complete list of the alternative instructions can be found in \cref{fig:questions} in the Appendix.

\subsection{Postprocessing} 
While most of the GPT-4 generated question-answer pairs are of high quality, some answers occasionally contain unneeded contents. For example, some answers may tell that the information is based on the given ``captions'' and ``descriptions''. 
To remove the unneeded content, we find them based on relative words and use GPT-4 to regenerate the responses. 
In addition, the number of generated conversations or QA pairs may be fewer than the requirement. We also remove them and generate new responses. 
We use the same procedure to filter the regenerated content until it is satisfying.

\subsection{Statistics and Analysis}
\paragraph{Statistics.} Employing the two-pass procedure, we obtain an extensive collection of data, including 1,565,797 conversation QAs, 1,556,902 complex reasoning QAs, 1,011,338 referring QAs and 106,274 detailed image descriptions. The averaging question and answer lengths are 9.6 and 27.9 words in \emph{conversation} subset, 12.6 and 26.6 words in \emph{complex reasoning} subset and 11.3 and 20.6 words in \emph{referring QAs} subset, respectively. In contrast, the mean length is 5.7 words per question and 1.8 words per answer in the original Visual Genome. The detailed descriptions in our dataset have 361.5 words on average, while the length of COCO dataset image captions is 11.3. Therefore, the corpus provided by our SVIT is of higher quality.

\paragraph{Distribution.} We analyze the distribution of question types in \emph{conversation}, \emph{complex reasoning} and \emph{referring QAs} tasks by visualizing the distribution of first three words in \cref{fig:distribution}. 
We can see that ``what'' questions are the largest category, in \emph{conversation} (38\%), \emph{complex reasoning} (55\%) and \emph{referring QAs} (41\%).
In the case of \emph{conversation}, question types are diverse, including simple yes-no questions, questions on object details, conditions and functions, etc. 
Regarding \emph{complex reasoning}, since we explicitly prompt GPT-4 to generate questions that need complex reasoning to answer, we collect a larger proportion of complex questions that commence with ``why'' (9\%) and ``how'' (11\%). Furthermore, most questions starting with ``how'' are simple object counting questions, i.e. ``how many'', in existing visual question answering datasets such as Visual Genome \cite{VisualGenome} and VQA \cite{goyal2017making}, while in SVIT, only 11\% of questions starting with ``how'' are the ``how many'' questions.
For \emph{referring QAs}, there are various types of questions, including those about object positions that start with ``where'', about object existence that start with ``is/are there any'', about suggestions and planning that start with ``what suggestion'' and about reasoning that start with ``why''. To better distinguish objects in the same image, there is also a notable proportion of questions that starts with ``which''.

\begin{figure*}
\centering
    \subfigure[Wrong caption in COCO dataset: ``\underline{Three} men and one older woman stand near a man who is looking in the mirror with the collar of his white shirt up.'']{
        \begin{minipage}[t]{.47\textwidth}
             \centering
             \includegraphics[height=3.5cm]{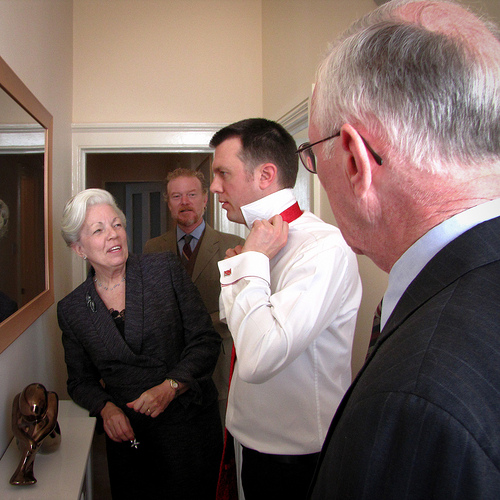}
             \label{fig:errors1}
        \end{minipage}
    }
    \subfigure[Wrong object name in Visual Genome: ``\underline{teddy bear}''.]{
        \begin{minipage}[t]{.47\textwidth}
            \centering
            \includegraphics[height=3.5cm]{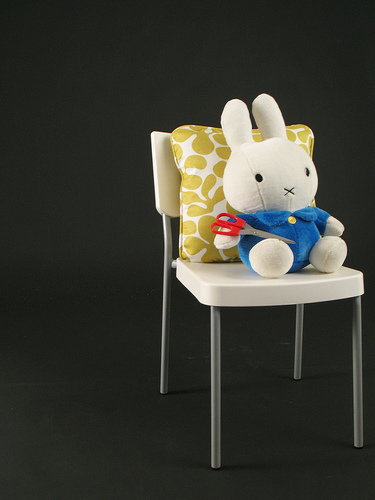}
            \label{fig:errors2}
        \end{minipage}
    }
    \\
    \subfigure[The answer discusses how the condition of the boat's paint would reflect the maintenance instead of answering it directly.]{
        \begin{minipage}[t]{.47\textwidth}
            \centering
            \includegraphics[height=3.5cm]{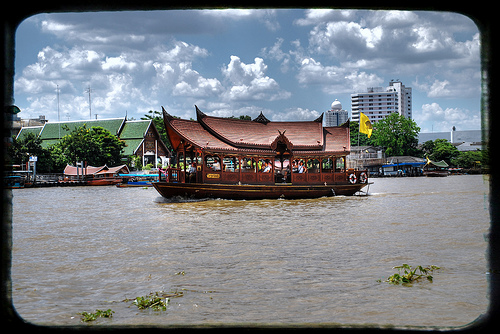}
            \label{fig:errors3}
        \end{minipage}
    }
    \quad
    \subfigure[The generated answer misunderstands the position of the telephone.]{
        \begin{minipage}[t]{.47\textwidth}
            \centering
            \includegraphics[height=3.5cm]{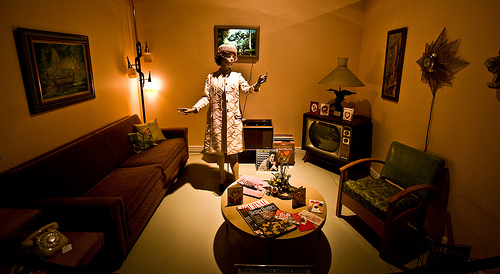}
            \label{fig:errors4}
        \end{minipage}
    }
    \caption{Problematic examples in generated answers.}
    \label{fig:errors}
\end{figure*}


\paragraph{Correctness.} To assess the correctness of the generated content, we conduct a manual examination on randomly selected 20 images and corresponding data. In general, around 5\% of the questions in the dataset can be provided with a more accurate or satisfying answer.
The identified problems can be categorized into three types.

\begin{itemize}
    \item Errors in original annotations. We construct the visual instruction data based on the manual annotations from Visual Genome and COCO dataset, which may contain errors in their original annotations. For example, in the image depicted in \cref{fig:errors1}, one caption from COCO dataset incorrectly states, ``Three men and one older woman stand near a man who is looking in the mirror with the collar of his white shirt up.'' Actually, there are only two men and one woman standing near the man looking at the mirror. Similarly, in \cref{fig:errors2}, the object is labeled as a ``little bunny'' in the region description, but wrongly referred to a ``teddy bear'' in the object name in Visual Genome's annotation.
    \item Correct but not precisely answer the question. As illustrated in \cref{fig:errors3}, when being asked, ``What can be inferred about the maintenance of the boat from the condition of the paint?'', the answer states, ``The condition of the boat's paint could reflect the level of maintenance, if it's faded or peeling, it may suggest the boat hasn't been maintained well, whereas bright and fresh paint may indicate regular upkeep.'' Although the answer is correct, it fails to address the question precisely.
    \item Incorrect answers. In \cref{fig:errors4}, the generated image description mentions, ``Nearby, there's a round center table cluttered with assorted magazines and books, creating a lived-in feel. The table also hosts a yellow rotary telephone, a vintage relic of bygone days.'' In reality, there are two tables in the image and the telephone is placed on a different table in the bottom left corner, though it needs careful observation.
\end{itemize}

%% file: method.tex
\section{Method}
\subsection{Model Architecture}
We employ the open-source Multimodal Large Language Model - LLaVA \cite{LLaVA,llava1.5}, which consists of a vision encoder $\psiV(\cdot, \bthetaV)$, a large language model $\psiL(\cdot, \bthetaL)$ and a vision-language connector $\psiC(\cdot, \bthetaC)$. We illustrate the model in \cref{fig:model}.
Provided with the input image $\bxV$ and instruction $\bxI$, the vision encoder is utilized to extract the image features $\bm{f} = \psiV(\bxV, \bthetaV)$.
Then a vision-language connector is applied to convert the image features to the language embedding tokens $\psiC(\bm{f}, \bthetaC)$. 
After that, the vision and language tokens are combined and fed into the LLM to generate the response:
\begin{equation}
    \Tilde{\bxR} = \psiL([\psiC(\bm{f}, \bthetaC), \bxI], \bthetaL)
\end{equation}

{
The training procedure contains two stages, including the pre-training on image-text pairs and fine-tuning on visual instruction data. In the pre-training stage, the vision-language connector parameters are updated using image-text pairs, while the weights of vision encoder and LLM remain frozen. 
In the fine-tuning stage, we implement full-parameter tuning or Low-rank Adaption (LoRA) tuning \cite{lora}.
Without ambiguity, $\bthetaL$ denotes the LLM parameters in full training setting and the learnable LoRA parameters in LoRA training setting. Then, the connector and learnable LLM parameters are updated using visual instruction data:}
\begin{equation}
    \bthetaC^*, \; \bthetaL^* = \argmin_{\bthetaC, \; \bthetaL} -\sum_{i=1}^N\sum_{j=1}^L \log p(\Tilde{\bxR}_i^j|{\bxV}_i, {\bxI}_i, {\bxR}^{<j}_{i}),
\end{equation}
where $N$ and $L$ denote the training sample size and the length of each response.



\subsection{Coreset Selection Algorithm}
The popular benchmarks evaluate different abilities of Multimodal Large Language Models (MLLM), which require specific recipe of training data to evoke the pre-trained model. Thus, we design a new data recipe, i.e. a coreset selection algorithm, to better adapt those benchmarks and achieve balance between performance and training efficiency. 

\paragraph{Diversity.} {We construct a set of key concepts that match the popular benchmarks, namely, MME \cite{mme} and MMBench \cite{mmbench}.} Specifically, we design several high-level concepts and then use GPT-4 to generate dozens of key words about each concept. Then, we filter out those key words that have low frequency in SVIT dataset. The concept set is illustrated in \cref{tab:dict} in the Appendix. We measure the informativeness of each training sample by its overlap with concept set, and select the most informative ones.

\paragraph{Balance.} ``Yes'' or ``No'' questions are used to evaluate models in MME benchmark. However, the proportion of the two choices in GPT-4 generated data is extremely unbalanced, which makes the tuned model has tendency to respond ``Yes''. We adjust the proportion by re-sampling. {We empirically study the relation between ``Yes:No'' proportion and model performance in \cref{sec:Y-N-balance}}.

With the above two operations, we obtain the coreset \textbf{SVIT-core-150K} of 157,712 samples, which has the same size as LLaVA-Instruct-150K. We also produce \textbf{SVIT-mix-665K} by replacing LLaVA-Instruct-150K in LLaVA-v1.5-mix-665K \cite{llava1.5} with SVIT-core-150K. 

%% file: experiments.tex
\section{Experiments}
Firstly, we compare our model to the state-of-the-art MLLMs in \cref{sec:sota}. In this sub-section, we tune the advanced LLaVA-v1.5-13B \cite{llava1.5} on the constructed SVIT-mix-665K dataset. Secondly, we implement ablation study and provide more detailed evaluations in \cref{sec:ablationstudy}. We tune the LLaVA-v1.0 (LLaVA-LLaMA-2-7B-Chat) \cite{LLaVA} with various data recipes for efficiency. Lastly, qualitative evaluation is provided in \cref{sec:qualitative}.

\begin{table*}[ht]
\centering
\caption{Comparison to state-of-the-art MLLMs on 11 benchmarks. Our models outperform LLaVA-v1.5 and others in most of the settings. 
We evaluate these models on benchmarks: VQA-v2 \cite{goyal2017making} test-dev split, GQA \cite{GQA2019} test-dev-balanced split, VisWiz \cite{gurari2018vizwiz} test-dev split, SQA$^\text{I}$: ScienceQA-IMG \cite{lu2022learn} test split, VQA$^\text{T}$: TextVQA \cite{singh2019towards} validation split, MME$^\text{P}$: MME perception \cite{mme}, MME$^\text{C}$: MME cognition \cite{mme}, MMB: MMBench \cite{mmbench} test split, MMB$^\text{CN}$: MMBench-Chinese \cite{mmbench} test split, SEED: SEED-Bench \cite{seedbench}, and MMMU \cite{yue2023mmmu} test split. 
We mark the best performance \textbf{bold} and the runner-up \underline{underlined}. 
$^*$The training images of the datasets are observed during training. 
$^\mathsection$We evaluate the officially released checkpoint by ourselves.}
\vskip 0.15in
\resizebox{\linewidth}{!}{
\begin{tabular}{l l | p{8mm}p{6mm}p{8mm}p{7mm}p{8mm} | p{8mm}p{8mm}p{7mm}p{9mm}p{7mm}p{8mm} }
\toprule
Method & LLM & VQA$^\text{v2}$ & GQA & VisWiz & SQA$^\text{I}$ & VQA$^\text{T}$ & MME$^\text{P}$ & MME$^\text{C}$ & MMB & MMB$^\text{CN}$ & SEED & MMMU \\
\midrule
BLIP-2 &  Vicuna-13B & -- & 41.0 & 19.6 & 61.0 & 42.5 & -- & -- & -- & -- & -- & -- \\
BLIP-2 &  Flan-T5-XXL & 65.0 & 44.6 & 29.4 & 64.5 & 44.1 & 1293.8 & 290.0 & -- & -- & -- & \underline{34.0} \\
InstructBLIP & Vicuna-7B & -- & 49.2 & 34.5 & 60.5 & 50.1 & -- & -- & 33.9 & 23.9 & 53.4 & -- \\
InstructBLIP & Vicuna-13B & -- & 49.5 & 33.4 & 63.1 & 50.7 & -- & -- & -- & -- & -- & -- \\
InstructBLIP &  Flan-T5-XXL & -- & 47.9 & 30.9 & 70.6 & 46.6 & -- & -- & -- & -- & -- & 33.8 \\
Shikra-7B & Vicuna-7B & -- & -- & -- & -- & -- & -- & -- & 60.2 & -- & -- & -- \\
Shikra-13B & Vicuna-13B & 77.4$^*$ & -- & -- & -- & -- & -- & -- & -- & -- & -- & -- \\
IDEFICS-9B & LLaMA-7B & 50.9 & -- & 35.5 & 44.2 & 25.9 & -- & -- & 45.3 & 25.2 & -- & -- \\
IDEFICS-80B & LLaMA-65B & 60.0 & -- & 36.0 & 68.9 & 30.9 & -- & -- & 54.6 & 38.1 & -- & -- \\
Qwen-VL & Qwen-7B & {79.5}$^{*}$ & 59.3$^{*}$ & 35.2 & 67.1 & \textbf{63.8}$^{*}$ & -- & -- & 32.2 & 7.8 & 56.3 & -- \\
Qwen-VL-Chat & Qwen-7B & 78.2$^{*}$ & 57.5$^{*}$ & 38.9 & {68.2} & \underline{61.5}$^{*}$ & 1487.6 & \underline{360.7} & 61.8 & 56.3 & {58.2} & 32.9 \\
mPLUG-Owl2 & LLaMA2-7B & 79.4$^{*}$ & 56.1$^{*}$ & 54.5 & 68.7 & 58.2 & 1450.2 & 313.2 & 66.0 & 60.3 & 57.8 & 32.1  \\
\midrule
LLaVA-v1.5 (LoRA) & Vicuna-13B & {80.0}$^{*}$ & {63.3}$^{*}$ & \textbf{58.9} & \underline{71.2} & {60.2} & {1541.7} & 300.4$^\mathsection$ & \underline{68.4}$^\mathsection$ & {62.4}$^\mathsection$ & {61.3} & {33.2}$^\mathsection$ \\
LLaVA-v1.5 (Full) & Vicuna-13B & {80.0}$^{*}$ & {63.3}$^{*}$ & {53.6} & \textbf{71.6} & {61.3} & {1531.3} & 295.4 & 67.8 & \textbf{63.3} & {61.6} & 33.6 \\ \midrule
SVIT-v1.5 (LoRA) & Vicuna-13B & \underline{80.1}$^{*}$ & \underline{63.4}$^{*}$ & \underline{56.7} & 69.9 & 61.1 & \underline{1560.3} &  \textbf{364.3} & 68.3 & \underline{63.2} & \underline{61.8} & \textbf{34.1} \\
SVIT-v1.5 (Full) & Vicuna-13B & \textbf{80.3}$^{*}$ & \textbf{64.1}$^{*}$ & 56.4 & 70.0 & 60.8 & \textbf{1565.8} & 323.2 & \textbf{69.1} & 63.1 & \textbf{61.9} & 33.3 \\
\bottomrule
\end{tabular}
}
\label{tab:sota-results}
\end{table*}

\subsection{Comparison to the State of the Art}
\label{sec:sota} 
We adopt LLaVA-v1.5-13B \cite{llava1.5} architecture and pre-training weights, and then tune it on the constructed SVIT-mix-665K, which is named \textbf{SVIT-v1.5}. Specifically, we replace LLaVA-v1.5-mix-665K with our SVIT-mix-665K in the visual instruction tuning stage. The rest of model training protocol is kept unchanged for fair comparison. 
Visual instruction tuning takes about 21 hours for both full-parameter tuning and LoRA tuning on 8 NVIDIA Tesla A100 GPUs, each with 80GB memory, with DeepSpeed ZeRO Stage 3.
We compare SVIT-v1.5 to state-of-the-art MLLMs: BLIP-2 \cite{Blip-2}, InstructBLIP \cite{instructblip}, Shikra \cite{shikra}, IDEFICS \cite{idefics}, Qwen-VL(-Chat) \cite{bai2023qwen}, mPLUG-Owl2 \cite{ye2023mplugowl2} and LLaVA-v1.5 \cite{llava1.5}. 
We evaluate these models on popular benchmarks: VQA-v2 \cite{goyal2017making}, GQA \cite{GQA2019}, VisWiz \cite{gurari2018vizwiz}, ScienceQA-IMG \cite{lu2022learn}, TextVQA \cite{singh2019towards}, MME perception \cite{mme}, MME cognition \cite{mme}, MMBench \cite{mmbench}, MMBench-Chinese \cite{mmbench}, SEED-Bench \cite{seedbench} and MMMU \cite{yue2023mmmu}. 

As shown in \cref{tab:sota-results}, our SVIT-v1.5 outperforms LLaVA-v1.5 and other models in most settings. Especially, in the most popular benchmark - MME, SVIT-v1.5 (Full) achieves 1565.8 score in MME perception and overwhelms LLaVA-v1.5 (Full) by 34.5 score. In the efficient LoRA training setting, SVIT-v1.5 (LoRA) exceeds LLaVA-v1.5 (LoRA) by 63.9 score, namely, 364.3 v.s. 300.4, in MME cognition. The improvements verify the better training effects of SVIT data, since the same data amount and base model are used.

\begin{table*}[ht]
\vspace{-5pt}
    \centering
    \caption{Evaluating models fine-tuned on LLaVA-Instruct-80K, SVIT-80K (random selection), {SVIT-80K-D (enhancing diversity)}, SVIT-80K-B (with ``Yes/No'' balancing) and SVIT-train (SVIT train split) on MME benchmark. Note that the base model is {LLaVA-v1.0} \cite{LLaVA}. For LLaVA-Instruct-80K, we evaluate the officially released checkpoint by ourselves.}
    \vskip 0.15in
    \resizebox{\linewidth}{!}{
    \begin{tabular}{ccccccccccc}
        \toprule
        Task & Sub-task & LLaVA-Instruct-80K & SVIT-80K & SVIT-80K-D &SVIT-80K-B & SVIT-train\\
        \midrule
        Overall & Total & 1147.70 & 1241.84& 1262.15&1329.77  & 1399.66 \\
        \midrule
        \multirow{11}{*}{Perception} & Total & 906.63 & 1005.41& 1017.15&1035.13  & 1166.45 \\
        \cline{2-7}
        & Existence & 90.00 & 90.00&95.00 & 120.00 & 185.00 \\
        & Count & 55.00 & 115.00& 110.00&118.33 & 131.67 \\
        & Position & 56.67 & 53.33& 58.33& 58.33 & 56.67 \\
        & Color & 50.00 &50.00 & 55.00&58.33  & 100.00 \\
        & Posters & 116.33 & 143.20& 146.26&133.67  & 134.01 \\
        & Celebrity & 85.88 &75.88 &77.06& 84.71  & 77.35 \\
        & Scene & 152.75 &161.25 & 159.50& 153.75 & 153.25 \\
        & Landmark & 130.75 &148.25 & 151.00& 137.75 & 144.50 \\
        & Artwork & 96.75 & 111.00& 107.50&105.25  & 104.00 \\
        & OCR & 72.50 &57.50 &57.50&  65.00 & 80.00 \\
        \midrule
        \multirow{5}{*}{Cognition} & Total & 241.07 & 236.43 &245.00& 294.64 & 233.21 \\
        \cline{2-7}
        & Commonsense reasoning & 83.57 &86.43& 80.00& 87.14  & 80.71 \\
        & Numerical calculation & 45.00 &57.50 &55.00& 57.50  & 47.50 \\
        & Text translation & 57.50 & 50.00& 65.00&97.50  & 50.00 \\
        & Code reasoning & 55.00 &42.50 &45.00&  52.50 & 55.50 \\
        \bottomrule
    \end{tabular}
    }
    \label{tab:mme}
\end{table*}

\subsection{Ablation Study}
\label{sec:ablationstudy}
{We further study the data quality, diversity strategy, balance strategy and scaling-up effects.} 10\% of the images are randomly sampled from SVIT as the held-out testing set for evaluation. The training split is denoted as \textbf{SVIT-train}. Note that, for saving the training cost, we implement ablation study with the LLaVA-v1.0 model and evaluate on MME benchmark.
We denote the LLaVA-v1.0 model trained on SVIT data as SVIT-v1.0.

\paragraph{Data Quality.} 
{LLaVA-v1.0} employs LLaVA-Instruct-80K as the visual instruction tuning data. To demonstrate the quality of SVIT, we construct a subset of SVIT-train at same scale of LLaVA-Instruct-80K and fine-tune {LLaVA-v1.0} by replacing LLaVA-Instruct-80K with the SVIT subset. Without loss of generality, the subset is constructed by randomly sampling 20K data from \emph{conversation}, \emph{complex reasoning}, \emph{referring QAs} and \emph{detail description}, leading to a subset of 80K data in total, denoted as \textbf{SVIT-80K}. We adopt the same training protocol and hyper-parameters as {LLaVA-v1.0}. The training takes less than 1 hour on 8 NVIDIA Tesla A100 GPUs, each with 40GB memory, with DeepSpeed ZeRO Stage 3.

The evaluation results on MME benchmark are shown in \cref{tab:mme}. The model fine-tuned on SVIT-80K achieves higher performance (+8.2\%) than the model fine-tined by LLaVA-Instruct-80K. Specially, the model fine-tuned on SVIT-80K outperforms on ``count'' (+109.1\%), ``posters''(+23.1\%), ``scene (+5.6\%)'', ``landmark (+13.4\%)'', ``artwork (+14.7\%)'' in perception tasks, as well as ``commense reasoning''(+3.4\%), ``numerical calculation'' (+27.8\%)  in cognition tasks.
The high performance of SVIT on those tasks can be due to the fact that SVIT dataset is constructed with more detailed manual annotations of the images, and the prompts for GPT-4 to generate QAs are carefully designed to cover a wide range of tasks, evoking the model to understand the images more accurately and comprehensively.


\begin{figure}[h]
  \centering
  \includegraphics[width=\columnwidth]{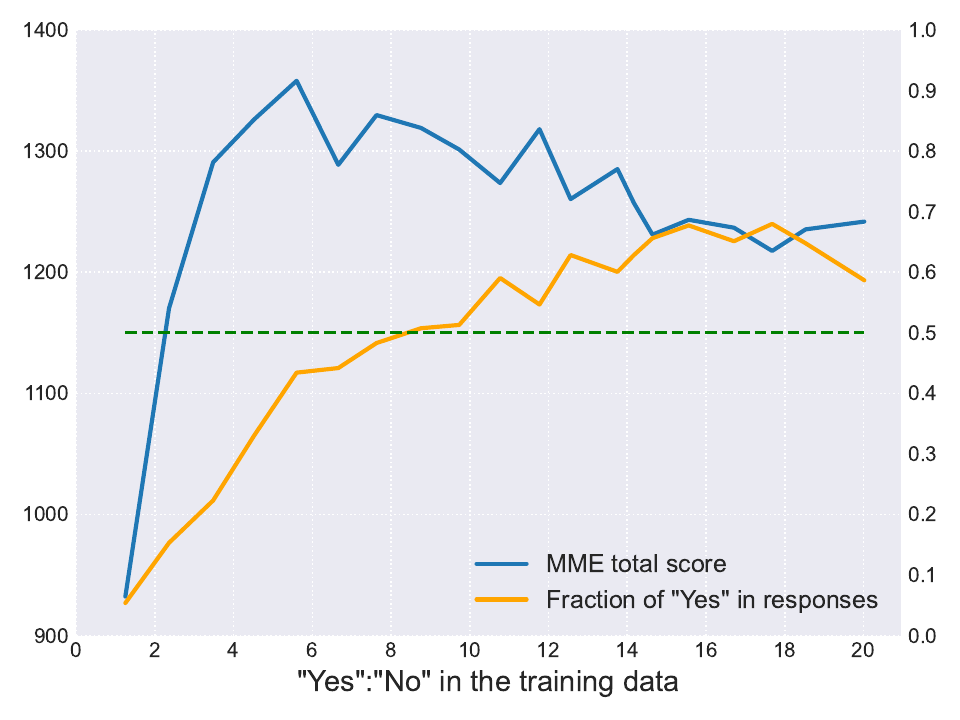}
  \vspace{-15pt}
  \caption{The relation between ``Yes:No'' proportion in training data and model performance.}
  \label{fig:llava-bias}
\end{figure}

\begin{figure*}[ht]
  \centering
  \includegraphics[width=\linewidth]{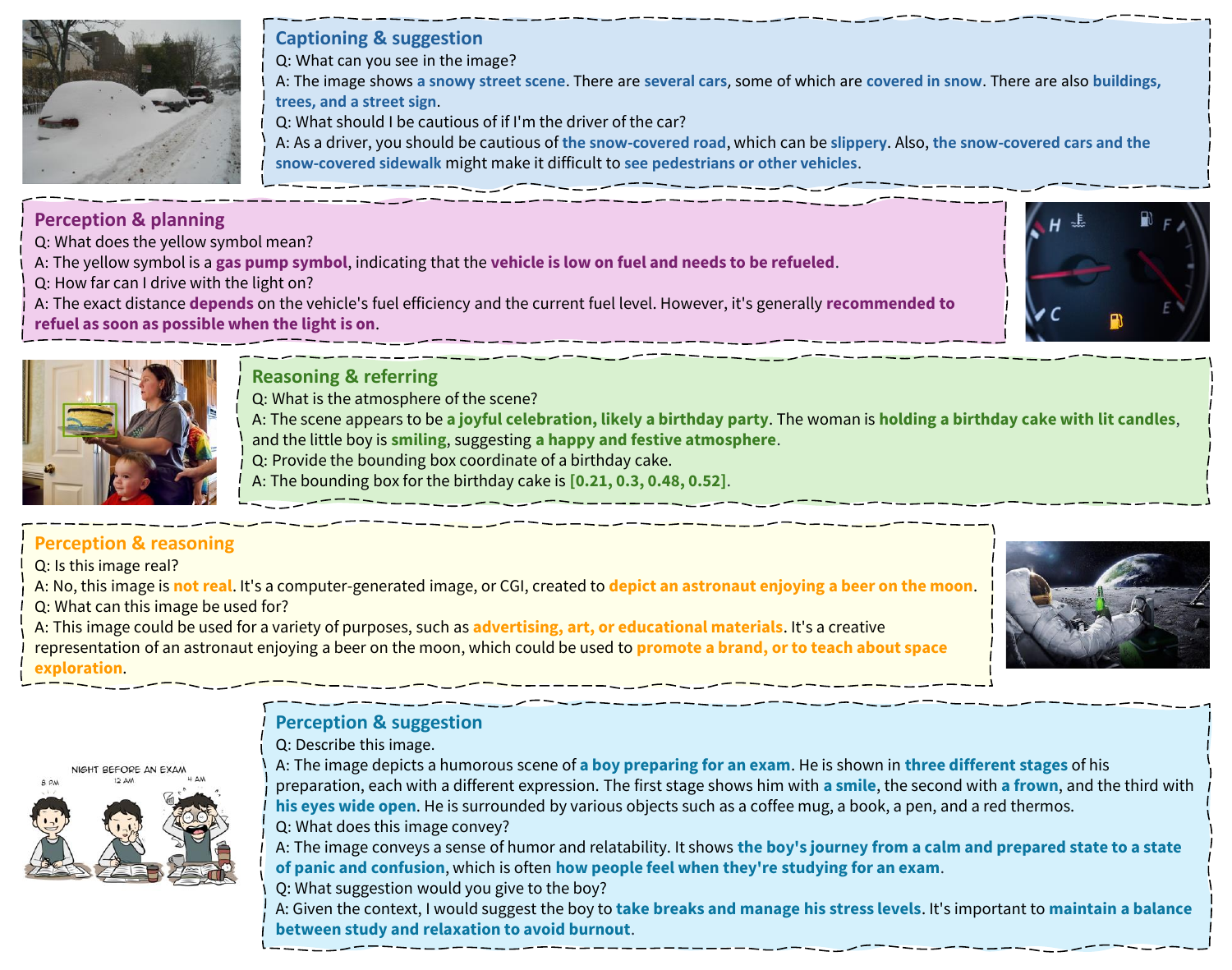}
  \caption{Demonstration of different abilities of SVIT-v1.5.}
  \label{fig:demo}
\end{figure*}

\paragraph{Diversity Strategy.} We produce an 80K subset selected with diversity strategy and compare it to the randomly selected SVIT-80K. We first remove half less-informative samples of SVIT-train based on the measured informativeness of each sample. Then we randomly sample 20K data for each category in SVIT, leading to the 80K subset -- \textbf{SVIT-80K-D}. As showed in \cref{tab:mme}, its performance has an improvement of 20.3 score over SVIT-80K, which verifies the effectiveness of the diversity strategy. 

\paragraph{Balance Strategy.} 
\label{sec:Y-N-balance}
{The MME benchmark consists of 2,374 ``Yes'' or ``No'' answers with the proportion $1:1$.} However, the randomly selected SVIT-80K dataset contains 7.5\% ``Yes'' or ``No'' QA pairs with the proportion $Y:N=20$. 
We analyze the relation between the ``Yes:No'' proportion and model performance by adjusting the proportion. 
As shown in \cref{fig:llava-bias}, the model trained on randomly sampled SVIT-80K with $Y:N=20$ responds 1,393 ``Yes'' while 981 ``No'' on MME questions.
We adjust the ``Yes:No'' proportion {in training data} by randomly dropping some questions with ``Yes'' answers {after random sampling from SVIT-train, ensuring a subset with exactly 80,000 samples.} It is interesting that the model trained on the equilibrium, i.e. $Y:N=1$, very likely responds ``No'' for any questions. The curve indicates that {$Y:N= 8$} is a good data recipe for {SVIT-v1.0 model} and the produced model will respond ``Yes'' or ``No'' uniformly, which is close to the prior. 
We denote the model tuned with this data recipe as \textbf{SVIT-80K-B}, {and it achieves 7.1\% improvement over the model fine-tuned on SVIT-80K on MME benchmark.}

\paragraph{Scaling Up.} 
To investigate whether scaling up the visual instruction tuning dataset actually helps improve the model's performance, we further conduct larger experiment - training the model with SVIT-train.
In the training process, the fine-tuning schedule and other hyper-parameters remain unchanged, while the learning rate is decreased from 2e-5 to 2e-6 to better fit the larger training data scale.
The training takes around 24 hours on 8 NVIDIA Tesla A100 GPUs, each with 40GB memory, with DeepSpeed ZeRO Stage 3.

The evaluation results are shown in the last columns of \cref{tab:mme}. We compare SVIT-train to SVIT-80K (randomly selected) without any data recipe. 
Compared with the model fine-tuned on SVIT-80K, the total score of the model fine-tuned on SVIT-train achieves +12.7\% score improvement on MME benchmark.
Particularly, fine-tuning the model on more data significantly enhances the model's ability to comprehend the existence of objects (+105.6\%), the color of objects (+100.0\%), OCR (+39.1\%), etc.
The results validate the effectiveness of scaling up the visual instruction tuning dataset when fine-tuning MLLMs.

\subsection{Qualitative Evaluation}
\label{sec:qualitative}
In \cref{fig:demo}, we provide the qualitative evaluation of the SVIT-v1.5.
The first case demonstrates a conversation discussing the scene and asking for suggestions. When describing the scene, SVIT-v1.5 depicts the foreground, as well as details in the background. When giving suggestions, SVIT-v1.5 offers a comprehensive assessment, taking multiple factors into consideration, such as the snow-covered road, cars and sidewalks.
In the second case, to evaluate the ability of planning, we ask the model what is happening in the image and prompt the model to plan the subsequent steps. SVIT-v1.5 accurately points out the meaning of the symbol and logically make recommendations.
In terms of the ability to locate and refer objects, the third case shows that SVIT-v1.5 correctly identifies the location of the mentioned object with a bounding box, in the format of [x1, y1, x2, y2], where [x1, y1] are the normalized coordinates of the top-left point and [x2, y2] are the normalized coordinates of the bottom-right point.
Regarding perception and reasoning performance, in the fourth case, SVIT-v1.5 is able to distinguish between a real image and a synthetic one. It also understands the intended use of the image, such as advertising, art or education.
Similarly, the fifth case feeds SVIT-v1.5 with comics about exam preparation. SVIT-v1.5 figures out that the three sub-figures are the different stages of the preparation and infers the theme of the comics. It also generates appropriate suggestions for the character.



%% file: conclusion.tex
\section{Conclusion}
In this paper, we scale up visual instruction tuning by presenting a large-scale dataset -- SVIT that contains in total 4.2 million instruction tuning data. We also propose new data recipe of sample selection for better diversity and balance.  
The abundant experiments verify that our SVIT-v1.5 trained on the proposed dataset and its subsets {outperforms} state-of-the-art MLLMs on multiple benchmarks. 

%% file: supplementary.tex








\icmltitle{Appendix}
\section{Prompts}

Based on the captions, object bounding boxes and region descriptions of images, we design four tasks and prompt GPT-4 to respond accordingly. We do not include the bounding boxes of region descriptions in the input data for \emph{conversation}, \emph{complex reasoning} and \emph{detail description}, since the context length may exceed the limit of GPT-4 in many cases. The prompts share the same paragraph describing the input data at the beginning and then differ in task description, which are summarized in \cref{fig:prompts}. For \emph{referring QAs}, since the location information plays a vital role in understanding the image accurately, we include the bounding boxes of region descriptions in the input data, and shorten the response to 10 QAs for every image to fit in the context limit. The prompt is summarized in \cref{fig:region-prompt}.

\begin{figure*}
\begin{tikzpicture}
    \node (example-textwidth-3) [draw, rounded corners,
                                 text width=\linewidth-24pt,    
                                 inner sep=12 pt]%
    { 
    \justify
    \small
You are an AI visual assistant that can analyze a single image. The information of the image is made up of three parts:
     
(1) ``captions'': If it is not empty, it contains five sentences, describing the image you are observing.

(2) ``objects'': It contains multiple lines, each describing an object of the same image you are observing. Every line is made up of an object name and its bounding box. The bounding box is in the form of [x1, y1, x2, y2]. The values are float numbers normalized from 0 to 1, corresponding to the top left x, top left y, bottom right x, and bottom right y.

(3) ``regions''. It contains multiple lines, each describing a region of the same image you are observing.

\hfill

\textbf{Conversation:}\par

Design three conversations between you and a person asking about this image. A conversation should include five question and answer pairs. The content within the conversation should be logically connected. You can think about the topic first and then generate the conversation according to the topic. The topic can be the visual content of the image (including the object types, counting the objects, object actions, object locations, relative positions between objects, etc.), the background knowledge of the objects, the events happening in the image, event planning, etc. In the conversation, you are called ``GPT''. The person talking with you is called ``User''.

\hfill

Ask diverse questions and give corresponding answers. Only include questions that have definite answers. The answer should be in a tone that a visual AI assistant is seeing the image and answering the question. The length of the answer would better be within 50 tokens.

\hfill

When using the information from the description, do not mention that the information source is the description. When using the information from the object bounding box, do not mention that the information comes from the bounding box as well. Always answer as if you are directly looking at the image.

\hfill

\textbf{Complex reasoning:}\par

Create 15 plausible question and answer pairs about the image with provided information.

\hfill

The question requires commonsense knowledge about the scene and can only be answered with image provided. Avoid asking questions that can be answered with commonsense knowledge alone. Avoid proposing questions that can be answered with simple visual understanding like asking about object type and color. Do not give too many details about the visual content of the image, so one has to figure it out first to answer the question correctly. The question can be asking why things happen that way, suggestions to the people in the image, etc. When providing the answer for complex questions, think step by step and include reasoning details.

\hfill

When using the information from the description, do not mention that the information source is the description. When using the information from the object bounding box, do not mention that the information comes from the bounding box as well. Always answer as if you are directly looking at the image.

\hfill

Desired format:

Question: ...

Answer: ...

Question: ...

Answer: ...

\hfill

\textbf{Detail description:}\par

The task is describing the image in detail. Though you do not receive the pixel data of the image, utilize above textual information to think about the image and describe as if you are directly looking at the image. The description can include what people or objects are doing, object appearance, object count, object position, background details, etc. Only describe the things that you are sure about.

\hfill

When using the information from the description, do not mention that the information source is the description. When using the information from the object bounding box, do not mention that the information comes from the bounding box as well.

    };
\end{tikzpicture}
\caption{The prompts of conversation, complex reasoning and detail description to GPT-4.}
\label{fig:prompts}
\end{figure*}

\begin{figure}
\begin{tikzpicture}
    \node (example-textwidth-3) [draw, rounded corners,
                                 text width=\linewidth-24pt,    
                                 inner sep=12 pt]%
    { 
    \justify
    \small
\textbf{Referring QAs:}\par

You are an AI visual assistant that can analyze a single image. The information of the image is made up of three parts:
(1) ``captions'': If it is not empty, it contains five sentences, describing the image you are observing.

(2) ``objects'': It contains multiple lines, each describing an object of the same image you are observing. Every line is made up of the object name and its bounding box.

(3) ``regions'': It contains multiple lines, each describing a region of the same image you are observing. Every line is made up of the region description and the region's bounding box.

\hfill

The bounding box is in the form of [x1, y1, x2, y2]. The values are float numbers normalized from 0 to 1, corresponding to the top left x, top left y, bottom right x, and bottom right y. Note that the same object may be described multiple times in ``objects'' with bounding boxes that are slightly different.

\hfill

The task is creating 10 question and answer pairs of specific regions in the image with the provided information. The question can only be answered with image information provided. Figure out the relative positions of objects and create some questions about that. Also propose some questions that need reasoning, like why things happen that way, suggestions to the people in the image, etc. When providing answers for complex questions, think step by step and include reasoning details.

\hfill

When referring to any object in the question or answer, always wrap it with prefix ``$<$st$>$'', suffix ``$<$ed$>$'' and attach its bounding box after it, in the format of ``$<$st$>$man$<$ed$>$ [x1, y1, x2, y2]''. If multiple objects are referred to, attach all the corresponding bounding boxes after them, e.g., ``$<$st$>$men$<$ed$>$ [x1, y1, x2, y2], [x1, y1, x2, y2]''. Refer to at least one object in the question and answer.

\hfill

When using the information from the description, do not mention that the information source is the description. When using the information from the bounding box, do not mention that the information comes from the bounding box as well. Always answer as if you are directly looking at the image.

\hfill

Desired format:

Question: ...

Answer: ...

Question: ...

Answer: ...

    };
\end{tikzpicture}
\caption{The prompt of referring QAs to GPT-4.}
\label{fig:region-prompt}
\end{figure}

\section{Instructions for Detail Description}

\cref{fig:questions} shows the instructions for detail description. We prompt GPT-4 to generate different ways of saying ``can you describe the image in detail'' and accumulate all the instructions. For each image, we randomly sample one from the list as instruction.


\begin{figure}
\begin{tikzpicture}
    \node (example-textwidth-3) [draw, rounded corners,
                                 text width=\linewidth-24pt,    
                                 align=flush center]%
    {\textbf{Instructions for detail description}\par 
    \small
    \begin{itemize}
        \setlength\itemsep{2pt}
        \item Can you provide a comprehensive description of the image?
        \item Elaborate on the details of the image.
        \item What are the specifics visible in the image?
        \item Could you offer an in-depth analysis of the image?
        \item Can you depict the image with precise detail?
        \item Give a detailed account of the image.
        \item Explain the image in meticulous detail.
        \item Can you portray the image in words?
        \item Give a thorough narrative of the image.
        \item Please provide an intricate breakdown of the image.
        \item Offer a complete interpretation of the image.
        \item Delve into the particulars of the image.
        \item Explain all the nuances you observe in the image.
        \item Provide a detailed commentary on the image.
        \item Illustrate the image in depth using your words.
        \item Could you give a blow-by-blow description of the image?
        \item Go into detail about the different elements of the image.
        \item Can you dissect the image and describe each element in detail?
        \item Detail the contents of the image extensively.
        \item Can you provide an in-depth explanation of the image?
        \item Provide a comprehensive overview of the image.
        \item Break down the elements of the image in detail.
        \item Can you expound upon the features of the image?
        \item Offer an exhaustive description of the image.
        \item How would you illustrate the image in words?
        \item Please convey the image's details verbally.
        \item Can you detail the contents of the image?
        \item Narrate what you see in the image in depth.
        \item Kindly provide a meticulous commentary on the image.
        \item Share an extensive description of the image.
        \item Could you interpret the image in a detailed manner?
        \item Present a detailed report of the image's features.
        \item Can you provide an intricate depiction of the image?
        \item Disclose every detail you see in the image.
    \end{itemize}
    };
\end{tikzpicture}
\caption{Instructions for detail description.}
\label{fig:questions}
\end{figure}

\section{Concept Set}
We design the concept set with the key words for measuring the informativeness of training samples, which is illustrated in \cref{tab:dict}. The key words for each concept are generated by prompting GPT-4 and filtering based on their frequencies occurring in the dataset. 

\begin{table*}[h]
    \centering
    \caption{The concept set and its key words for measuring the informativeness of training samples.}
    \vskip 0.15in
\begin{tabular}{l|l}
\toprule[2pt]
\textbf{Concept}      & \textbf{Key Words}      \\ \midrule[0.5pt]
color                  &  \multicolumn{1}{m{0.6\linewidth}}{beige, black, brown, color, gold, gray, green, khaki, lavender, mauve, olive, peach, pink, red, rose, salmon, white                                                               }                                                                                                                                                                                                            \\ \midrule[0.5pt]
material               &  \multicolumn{1}{m{0.6\linewidth}}{canvas, cardboard, ceramic, cork, denim, fabric, fiberglass, foam, glass, glassy, granite, iron, latex, leather, linen, marble, mesh, metal, nylon, plaster, plastic, polymer, porcelain, satiny, silk, steel, stone, stony, suede, velvet, vinyl, wood, wooden    }                                                                                                                                                                                                                                                         \\     \midrule[0.5pt]                                                                                                                 
 
quantity &  \multicolumn{1}{m{0.6\linewidth}}{account, being, existence, five, four, number, one, seven, six, substance, ten, three, total, two    }\\ \midrule[0.5pt]
spatial relation               &  \multicolumn{1}{m{0.6\linewidth}}{above, adjacent, ahead, backward, below, between, central, close, down, downward, far, in back, inside, left, left direction, near, on, outside, peripheral, position, proximate, remote, surrounding, under, up, upstairs, upward, without} \\ \midrule[0.5pt]
size                   &  \multicolumn{1}{m{0.6\linewidth}}{big, compact, compactness, dimension, diminutive, enormity, enormous, giant, gigantic, immense, immensity, large, largeness, magnitude, massive, medium size, microscopic, miniature, minuscule, moderately, oversized, proportion, sizeable, slightly, small, smaller, vast, vastness }                                                                                                       \\ 
\bottomrule[2pt]
\end{tabular}

    \label{tab:dict}
\end{table*}